\documentclass[letterpaper, 10 pt, conference]{ieeeconf}  

\IEEEoverridecommandlockouts                              

\usepackage{cite}
\usepackage{graphics} 
\usepackage{epsfig} 
\usepackage{amsmath} 
\usepackage{bm}
\usepackage{amssymb}  
\usepackage{algorithm}
\usepackage{algorithmic}
\usepackage{acronym}
\usepackage{hyperref}
\usepackage{xcolor}
\newcommand{\cmark}{\checkmark}
\newcommand{\xmark}{\ding{55}}
\usepackage{pifont}
\usepackage{multirow}
\usepackage{tikz}
\usepackage{booktabs}
\title{\LARGE \bf
SAC-Loco: Safe and Adjustable Compliant Quadrupedal Locomotion
}

\author{
  Aoqian Zhang\textsuperscript{1}\thanks{\textsuperscript{1}Department of Electrical and Computer Engineering, National University of Singapore, Singapore},
  Zixuan Zhuang\textsuperscript{2}\thanks{\textsuperscript{2}School of Computer Science and Engineering, Sun Yat-sen University, China},
  Chunzheng Wang\textsuperscript{1},
  Shuzhi Sam Ge\textsuperscript{1},
  Fan Shi\textsuperscript{1} and
  Cheng Xiang\textsuperscript{1}
}

\begin{document}

\maketitle
\begin{abstract}
Quadruped robots are designed to achieve agile and robust locomotion by drawing inspiration from legged animals. However, most existing control methods for quadruped robots lack a key capacity observed in animals: the ability to exhibit diverse compliance behaviors while ensuring stability when experiencing external forces. In particular, achieving adjustable compliance while maintaining robust safety under force disturbances remains a significant challenge. In this work, we propose a safety-aware compliant locomotion framework that integrates adjustable disturbance compliance with robust failure prevention. We first train a force-compliant policy with adjustable compliance levels using a teacher–student reinforcement learning framework, allowing deployment without explicit force sensing. To handle disturbances beyond the limits of compliant control, we develop a safety-oriented policy for rapid recovery and stabilization. Finally, we introduce a learned safety critic that monitors the robot's safety in real time and coordinates between compliant locomotion and recovery behaviors. Together, this framework enables quadruped robots to achieve smooth force compliance and robust safety under a wide range of external force disturbances.
\end{abstract}

\section{INTRODUCTION}
Compliance is a fundamental principle of locomotion observed in both humans and animals. When subjected to external forces, legged animals exhibit versatile responses: they may resist or yield to the disturbance. More importantly, under strong perturbations, animals instinctively move in the direction of the applied force to preserve balance and prevent falls. Achieving such safe and compliant behaviors in legged robots is essential for reliable real-world operations under external forces. Beyond self-protection, adjustable compliance could potentially broaden the rage of future applications, such as enhancing adaptability in complex environments, enabling safer human-robot interaction, and supporting cooperative tasks.

Reinforcement learning (RL) controllers have been shown to be highly effective in allowing quadruped robots to perform a wide range of tasks, such as complex terrain traversal \cite{cheng2024parkour}, physical interaction with objects \cite{ji2023dribblebot, arm2024pedipulate}, and obstacle avoidance \cite{he2024agile}. Although RL-based controllers enable diverse whole-body behaviors, achieving robust and adaptive response to external forces remains a significant challenge. Most existing approaches emphasize strict task execution, such as tracking velocity or position regulation \cite{kim2025highspeedcontrolnavigationquadrupedal}, \cite{zhang2024resilient}, while neglecting the effects of large or persistent external disturbances. Although deep RL inherently provides some robustness to perturbations, achieving versatile responses across a broader range of external forces requires more sophisticated learning strategies. To this end, both model-based approaches \cite{kang2024external} and learning-based methods \cite{xiao2024pa} have been explored for disturbance adaptation. However, these methods often produce limited behavioral diversity, leading to failures when facing large or prolonged disturbances. This highlights the need for a locomotion controller that can diversify its response to external disturbances through adjustable compliance, enabling it to resist or yield with applied forces while maintaining robust stability.

To address this challenge, we propose SAC-Loco, a safe and adjustable compliant quadrupedal controller that integrates a velocity-based force compliant controller with a safety-oriented recovery controller. A learned safety critic is proposed to detect and prevent potential failures promptly. The main contributions of this work are as follows:
\begin{itemize}
    \item An external force compliant locomotion policy trained with a teacher-student reinforcement learning framework, enabling a wide range of adjustable compliance behavior without requiring explicit force sensing.
    \item A safe policy grounded in the capture-point dynamics for robot recovery from dangerous states and stabilization under large force disturbance.
    \item A learned safety critic that evaluates the robot's recoverability, ensuring smooth compliance under safe interactions and timely recovery when destabilized by large disturbances.
    \item Extensive simulations and hardware experiments validate the effectiveness of the proposed method.
\end{itemize}

\begin{figure}[t]
\centering
\includegraphics[width=\linewidth]{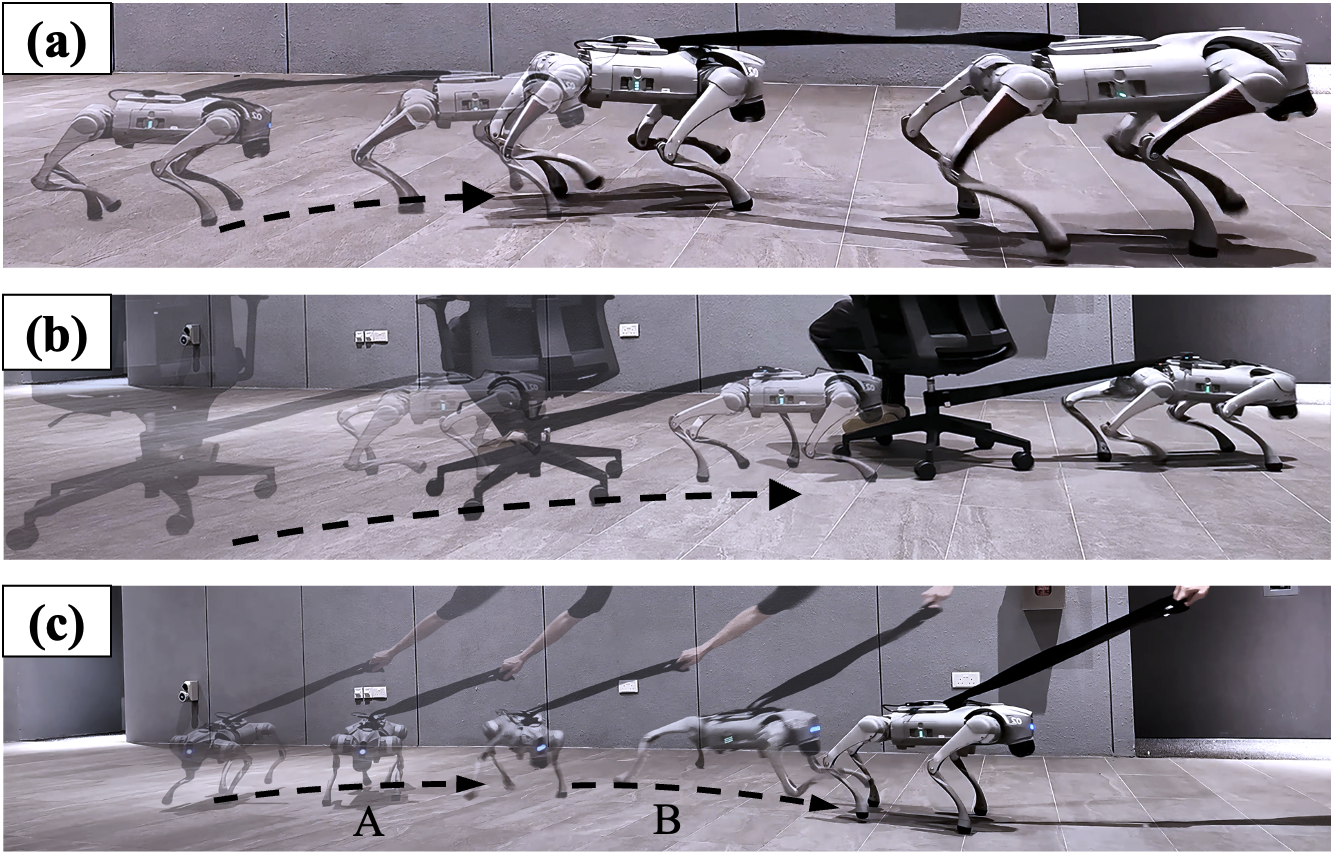}
\caption{(a) Quadruped robot compliantly follow the leader robot. (b) Quadruped robot pulling a human on a chair. (c) A: Quadruped robot move along with a pulling force compliantly and B: jump toward the pulling force to restore balance.}
\label{fig:summary}
\vspace{-0.7em}  
\end{figure}

\section{RELATED WORKS}
For external force–compliant and safe control of legged robots, some existing work adopts model-based methods to estimate the external forces acting on the robot. Khandelwal et al. propsed a method using only proprioceptive feedback to estimate the external forces \cite{khandelwal2025compliantcontrolquadrupedrobots}, which has been shown to provide force estimations that can be directly leveraged by an admittance controller for payload transportation. Beyond using proprioceptive information, Kang el al. propsed multimodal sensing by combining IMU and vision data to estimate external forces and moments through factor graph optimization \cite{kang2024external}. The estimated forces can then be incorporated into a model predictive control (MPC) framework to optimize footstep placement for disturbance rejection. Despite these advances, existing model-based approaches have two key limitations: (i) they are typically restricted to handling small perturbations (less than $50$N), and (ii) they rely on pre-defined gaits, which constrain the robot to low-speed locomotion. These limitations significantly undermine the robot’s ability to respond effectively to large, impulsive external forces. 

Reinforcement learning (RL) based methods, on the other hand, demonstrate stronger generalization and robustness to larger impulse disturbances. For example, Hartmann et al. proposed a multi-stage episodic training setup with reward shaping for robots to respond less aggressively to transient perturbations \cite{hartmann2024deep}. However, this approach is still not effective against large persistent forces. Inspired by biomechanics, Li et al. proposed a torque-based policy that directly outputs joint-level torques such that the quadruped robot can adapt to sudden disturbances acting on the legs and exhibits compliance to external forces in all directions \cite{li2025satasafeadaptivetorquebased}. However, the resulting fixed compliance prevents the robot from reliably following velocity commands under external pushes and doesn't incorporate adjustable compliance behavior. To address this limitation, Zhou el al. propsed to apply hierarchical reinforcement learning for active compliance control \cite{zhou2025hacloco}. In this framework, a high-level RL planner modulates velocity commands according to estimated external forces so that the robot can actively follow velocity targets while complying with external forces. Despite its advantages, the method has notable limitations: (i) the force estimation module is trained jointly with the low-level policy, which removes the necessity of the high-level RL planner, and (ii) modifying compliance parameters requires retraining the high-level policy, limiting flexibility and efficiency. Concurrently with our work, Xu el al. proposed a RL framework that integrates a virtual mass–spring–damper model for reference generation and tracking \cite{xu2025facet}. This approach enables robust velocity compliance with adjustable compliance levels. However, its success rate decreases when external forces exceed $500$N, highlighting the limitation of relying on a single policy to achieve both precise velocity tracking and high force tolerance. Our work, in comparison with existing methods, has serveral advantages as shown in Table \ref{Comparison}.

\begin{table}[t]
\vspace{5pt}
\centering
\caption{Comparison of different methods (M: Model-based, L: Learning-based). ES: No Extra sensor required, VC: Velocity compliance,  AC: Adjustable compliance level in a unified policy, LT: Large impulse toleration.}
\begin{tabular}{|l|c|c|c|c|}
\hline
\textbf{Method (paradigm)} & \textbf{ES} & \textbf{VC} & \textbf{AC} & \textbf{LT} \\
\hline
Kang et al. \cite{kang2024external} (M)& \xmark & \xmark & \xmark & \xmark \\
Hartmann et al. \cite{hartmann2024deep} (L)& \cmark & \xmark & \xmark &\xmark\\
Zhou et al. \cite{zhou2025hacloco} (L) & \cmark & \cmark & \xmark & \xmark \\
Xu et al. \cite{xu2025facet} (L) & \cmark & \cmark & \cmark & \xmark \\
\hline
\textbf{SAC-Loco (L)} & \cmark & \cmark & \cmark & \cmark \\
\hline
\end{tabular}
\label{Comparison}
\end{table}

\section{METHODS}
\begin{figure*}[h]
    \vspace{8pt}
    \centering
    \includegraphics[width=1.0\linewidth]{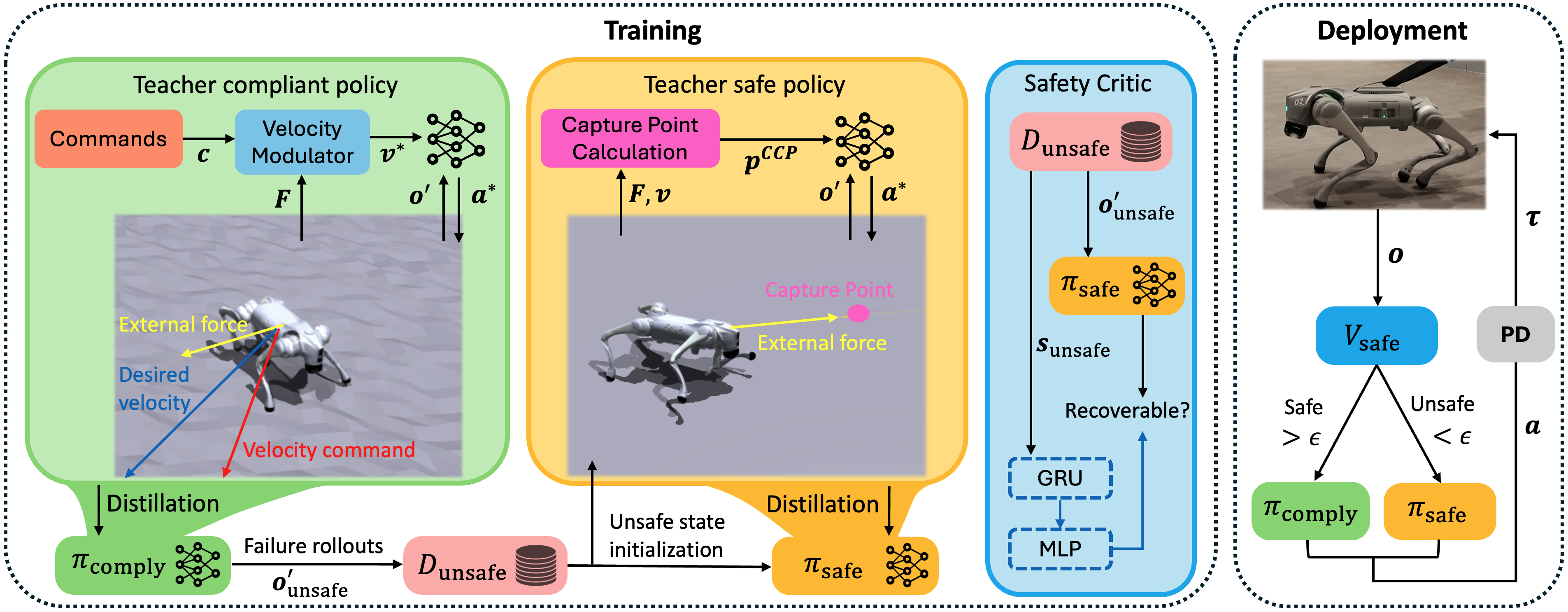}
    \caption{Overview of SAC-Loco. The teacher compliant policy $\pi_{\text{comply}}^*$ with velocity modulation is first trained using privileged observations. It is then distilled into a student compliance policy $\pi_{\text{comply}}$. The failure rollouts from $\pi_{\text{comply}}$ are collected into an unsafe dataset $\mathcal{D}_{\text{unsafe}}$. A teacher safe policy $\pi_{\text{safe}}^*$ is trained to recover from unsafe states initialized from $\mathcal{D}_{\text{unsafe}}$ and other large force disturbances using capture point dynamics. After distilling into the student safe policy $\pi_{\text{safe}}$, a safety critic $V_{\text{safe}}$ is trained to estimate the recoverability of $\pi_{\text{safe}}$. During deployment, $V_{\text{safe}}$ selects the policy to control the robot.}
    \label{fig:pipeline}
\vspace{-0.7em}   
\end{figure*}

We present SAC-Loco, a safety-aware locomotion control framework composed of three learned modules as illustrated in Fig. \ref{fig:pipeline}: (i) a compliant policy with adjustable compliance level under external disturbances (Section III-B), (ii) a safe policy designed to recover the robot from unsafe states and stabilize it under large disturbances (Section III-C), and (iii) a safety critic that evaluates the safety of the current state and activates the safe policy when recovery is required (Section III-D). 

\subsection{Problem Formulation}

We consider quadrupedal locomotion under external disturbances, where the robot receives a command $\bm{c} = [v_x', v_y', \omega_z', k]$ specifying a command linear velocity in robot's body frame, a command angular velocity about the $z$-axis and a tunable compliance level parameter. In the presence of an external force disturbance in robot's body frame $\bm{F} = [F_x, F_y, F_z]$, the control objective is to track the command velocity while exhibiting adjustable compliance to the external disturbances, yielding more to the force for larger values of $k$ and resisting more for smaller values of $k$. When the external disturbance becomes sufficiently large and may drive the robot into unsafe states, the control objective shifts to restore balance and stabilize the robot. In this work, a failure is defined as the quadruped robot's base making contact with the ground or the robot falling over. At each control step $t$, the control policies are executed on the quadruped robot using only proprioceptive sensing and generate joint-level actions. The observable states of the robot is given by  
\begin{equation}
\bm{s}_t = [\bm{g}_t,\bm{\omega}_t, \bm{q}_t, \bm{\dot{q}}_t, \bm{a}_{t-1}, \bm{c}_t],
\label{states}
\end{equation}
where $\bm{g}_t \in \mathbb{R}^3$ denotes projected gravity, $\bm{\omega}_t \in \mathbb{R}^3$ the body angular velocity, $\bm{q}_t, \dot{\bm{q}}_t \in \bm{R}^{12}$ the joint positions and velocities, $\bm{a}_{t-1} \in \mathbb{R}^{12}$ the previous actions, and $\bm{c}_t \in \mathbb{R}^{4}$ the command. The actions $\bm{a}_t\in \mathbb{R}^{12}$ of the control policies are the target joint positions $\bm{{q}}_t^*$, tracked by a low-level proportional-derivative (PD) controller that computes joint torques $\bm{\tau}_t = K_p \left( \bm{q^{*}}_t - \bm{q}_t \right) + K_d \left( -\bm{\dot{q}}_t \right)$, where $K_p$ and $K_d$ denote the proportional and drivative gains, respectively.

\subsection{Compliant Policy Training}
To achieve compliance behavior, a velocity modulator calculates the desired velocity $\bm{v}^* = [v_{x}^*, v_{y}^*]$ based on the external force 
\begin{align}
v_x^* &= \mathrm{clip}\!\left(v_x' + kF_x,\,-v_{x,\max},\,v_{x,\max}\right), \\
v_y^* &= \mathrm{clip}\!\left(v_y' + kF_y,\,-v_{y,\max},\,v_{y,\max}\right).
\label{vel_calculation}
\end{align}
The desired velocity, if followed by the robot, would enable the robot to achieve adjustable compliance behaviors under external force disturbances. 
To train a policy that tracks the desired velocity without external force information, we first train a teacher force compliant policy $\pi^*_{\text{comply}}$ using additional privileged observations obtained in simulations
\begin{equation}
\bm{o}_t' = [\bm{s}_t, z_t, \bm{Q}_t, \bm{v}_t, \bm{F}_t, \bm{\tau}_{\mathrm{ext},t}], 
\label{teacherobs}
\end{equation}
where $z_t$, $\bm{Q}_t \in \mathbb{R}^4$, $\bm{v}_t \in \mathbb{R}^3$, $\bm{F}_t \in \mathbb{R}^3$, and $\bm{\tau}_{\mathrm{ext},t} \in \mathbb{R}^3$ correspond to base height, body orientation in quaternions, body velocity, external force and external torque disturbances, respectively.
We design the velocity tracking rewards to enable the robot to follow the desired velocity $\bm{v}^*$ computed by the velocity modulator. Regularization rewards are included to help robot maintain stable posture and smooth joint level actions. The detailed reward terms and corresponding weights are shown in Table \ref{tab:reward_terms1}. 

After training $\pi^*_{\text{comply}}$, we distill it into student policies $\pi_{\text{comply}}$ that can be deployed on the physical robot using only the observable states defined in \eqref{states}. To compensate for the lack of privileged information, the student’s policy receives as input a stacked observation history of length 20
\begin{equation}
\bm{o}_t = [\bm{s}_{t-20}, \bm{s}_{t-19}, ..., \bm{s}_{t-1}, \bm{s}_{t}]. 
\label{studentobs}
\end{equation}
To achieve efficient training and a balance between exploration and exploitation, we use the PPO distillation method in \cite{zhang2025distillationppo}. We compute the mean squared error loss between the teacher's actions and the student's actions
\begin{equation}
\mathcal{L}_{\text{distillation}} = \left\| \pi^*{(\bm{o}_t')} - \pi(\bm{o}_t) \right\|_2^2,
\end{equation}
and the total loss for PPO becomes
\begin{equation}
\mathcal{L}_{\text{total}} 
= \alpha \mathcal{L}_{\text{distillation}} 
+ \beta \mathcal{L}_{\text{PPO}}.
\end{equation}
To facilitate early stage exploration in the correct direction, we also added two additional reward terms inspired by \cite{xu2025aclactionlearnerfaulttolerant} to guide the student policy's actions to stay close to the teacher policy's actions
\begin{equation}
\small
r_{\text{action}}=\exp\!\left(-\frac{\lVert \bm{a}_t^*-\bm{a}_t\rVert^2}{0.5}\right),\;
r_{\text{direction}}=\exp\!\left(\frac{\bm{a}_t\!\cdot\! \bm{a}_t^*}{\lVert \bm{a}_t\rVert\,\lVert \bm{a}_t^*\rVert}\right),
\end{equation}
where $\bm{a}^*_t$ and $\bm{a}_t$ are the teacher's and student's actions, respectively. To increase training efficiency, we adopt asymmetric PPO, where the critic additionally observes the priviledged observation $\bm{o}'_{t}$ in \eqref{teacherobs}. 

\subsection{Safe Policy Training}
The safe policy's objective is to prevent failure under large force disturbances by actively restoring balance of the robot. 
Based on the linear inverted pendulum model\cite{pratt2006capture}, the Corrected Capture Point (CCP) determines the required planar displacement of the support polygon centroid (SPC) of the quadruped robot to neutralize external forces\cite{kang2024external}. We calculated the position of SPC for a quadruped robot as 
\begin{equation}
\bm{p}^{\mathrm{cent}}_t =
\begin{cases}
\dfrac{1}{|\mathcal{S}_t|}
\sum\limits_{i \in \mathcal{S}_t} \bm{p}_{i,t},
& \text{if } |\mathcal{S}_t| \ge 2, \\[10pt]
\bm{p}^{\mathrm{base}}_t,
& \text{if } |\mathcal{S}_t| < 2,
\end{cases}
\label{eq:support_centroid_piecewise}
\end{equation}
where $\mathcal{S}_t \in \mathbb{R}^2$ is the set of supporting feet at time $t$, $\bm{p}_{i,t} \in \mathbb{R}^2$ is the planar position of foot $i$ and $\bm{p}^{\mathrm{base}}_t \in \mathbb{R}^2$ is the planar position of robot's base. To restore balance under external forces, $\bm{p}^{\mathrm{cent}}_t$ needs to be shifted to $\bm{p}^{\mathrm{CCP}}_t = \bm{p}^{\mathrm{cent}}_t + [\delta x_t, \delta y_t]^\top$ by the planar offsets
\begin{align}
    \delta x_t &= \sqrt\frac{z_t}{g} {v_{x,t}} + \frac{F_{x,t} z_t}{m g} \label{eq:ccpx}, \\
    \delta y_t &= \sqrt\frac{z_t}{g} {v_{y,t}} + \frac{F_{y,t} z_t}{m g} \label{eq:ccpy},
\end{align}
where $z_t$ is the base height of the robot, $m$ is the mass of the robot and $g$ is the gravitational constant. For a quadruped robot, due to its structural anisotropy, it can sustain a larger force in the longitudinal ($x$-axis) direction than in the lateral ($y$-axis) direction. Therefore, we further compute a yaw orientation offset
\begin{equation}
\delta\psi =
\begin{cases}
\operatorname{atan2}(F_y, F_x), & \text{if } F_x \geq 0 \quad , \\
\operatorname{atan2}(F_y, F_x) + \pi, & \text{if } F_x < 0 \quad,
\end{cases}
\label{eq:force_yaw}
\end{equation}
to align the robot's $x$-axis with the disturbance force. When the force acts toward the robot’s front, the robot aligns its heading with the force direction; when the force acts toward the rear, the robot aligns its tail with the force direction, thereby converting lateral disturbances into longitudinal ones. 

In order for the robot to adjust its $\bm{p}^{\mathrm{cent}}_t$ to $\bm{p}^{\mathrm{CCP}}_t$ as fast as possible, we train a teacher safe policy $\pi_{\text{safe}}^*$ in two stages. In Stage I (Pose tracking), we randomly sample target offset $\delta x, \delta y \sim \mathcal{U}(-3\,\mathrm{m},\, 3\,\mathrm{m})$ and $\delta\psi \sim \mathcal{U}(-\pi\,\mathrm{rad},\, \pi\,\mathrm{rad})$, training the robot to track any pose as fast as possible.
In Stage II (Disturbance recovery), we start exerting external forces on the robot and calculate $\delta{x}_t$, $\delta{y}_t$ and $\delta\psi_t$ at each timestep using \eqref{eq:ccpx} - \eqref{eq:force_yaw}. This policy takes the same observation as \eqref{teacherobs}, with $\bm{c}_t$ replaced by $\delta{x}_t$, $\delta{y}_t$ and $\delta\psi_t$.  We design reward functions to minimize the position tracking error ${p}^{err}_t
= \left\lVert \bm{p}^{\text{cent}}_t - \bm{p}^{\text{CCP}}_t \right\rVert_2$ between the robot's SPC and target CCP, and the yaw orientation offset $\delta \psi$. The detailed reward terms and their corresponding weight are shown in Table \ref{tab:reward_terms1}. After training $\pi_{\text{safe}}^*$, we distill it into the student safe policy $\pi_{\text{safe}}$ using the same distillation method in Section III-B.

\begin{table}
\vspace{5pt}
\centering
\caption{Primary Reward functions for $\pi_{\text{comply}}$ (marked $\dagger$), 
$\pi_{\text{safe}}$ (marked $\star$), shared regularization terms (no mark), 
and their corresponding weights.}
\renewcommand{\arraystretch}{1.5}
\resizebox{0.99\linewidth}{!}{
\begin{tabular}{l c c}
\hline
\textbf{Name} & \textbf{Expression} & \textbf{Weight} \\
\hline
Linear velocity tracking $\dagger$ & $
\exp\left( -\frac{\lVert v_{xy}^* - v_{xy} \rVert_2}{0.25} \right)
$& 1.0\\
Angular velocity tracking $\dagger$ &$
\exp\left( -\frac{( \omega_{z}' - \omega_z)^2}{0.25} \right)
$& 0.5\\
Position tracking soft $\star$& $\frac{1}{1 + {p^{err}}^2}$& 20.0\\
Position tracking tight $\star$&$\frac{1}{1 + ({p}^{err}/{0.25})^2}$& 20.0\\
Yaw angle tracking $\star$&$\frac{1}{1 + ( \delta \psi/0.5 )^2}$& 20.0\\
Velocity direction $\star$&${\max\!\left(0,\; \tfrac{v_x\delta x + v_y \delta y}{{p}^{err}} \right)}$& 40.0\\
Yaw rate direction $\star$&$\dot{\psi} \cdot 
\operatorname{sign}\!\big(\delta \psi)$&20.0\\
Stand still $\star$ & $\begin{aligned}
   & (\|v_{xy}\| < 0.5 \;\land\; |\dot{\psi}| < 0.1 )\cdot \mathbf{1}_{\text{reached}}
\end{aligned}$ & 20.0 \\
Base height&$(z-0.25)^2$ & -0.5\\
Angular velocity penalty&$\left\lVert \omega_{xy} \right\rVert^2$& -0.05\\
Joint torques&$\|\boldsymbol{\tau}\|_2^2$&-0.0005\\
Joint velocities&$\|\boldsymbol{\dot{\bm{q}}}\|_2^2$& -0.0001\\
Joint accelerations&$\left\lVert \frac{\dot{\bm{q}}_{t-1} - \bm{\dot{q}}_{t}}{dt} \right\rVert^2$&-2.5e-7\\
Action rate&$\left\lVert \frac{\dot{\bm{a}}_{t-1} - \dot{\bm{a}}_{t}}{dt} \right\rVert^2$&-0.01\\
Joint angle limit&$
\sum_{i=1}^{12} \!\left( \lvert q_i \rvert - 0.9 \cdot q_{i,\mathrm{lim}} \right)
$& -10.0\\
Joint velocity limit&$
\sum_{i=1}^{12} \!\left( \lvert \dot{q}_i \rvert - 0.9 \cdot \dot{q}_{i,\mathrm{lim}} \right)
$& -5.0\\
Joint torque limit&$
\sum_{i=1}^{12} \!\left( \lvert \tau_i \rvert - 0.85 \cdot \tau_{i,\mathrm{lim}} \right)
$& -5.0\\
Collision&$F_{\text{base, thigh, calf}}>0$& -10.0\\
\hline
\end{tabular}}
\label{tab:reward_terms1}
\vspace{-10pt}
\end{table}

\subsection{Safety Critic Training}
To enable timely switching from the compliant policy $\pi_{\text{comply}}$ to the safe policy $\pi_{\text{safe}}$ before failure occurs, we train a safety critic $V_{\text{safe}}$ that evaluates the robot's observations in real time. Unlike prior policy switching approaches (e.g.\cite{he2024agile}, \cite{jain2019hierarchicalreinforcementlearningquadruped}), which rely on fixed policy switching rules or only condition on a single policy, our proposed critic leverages information from both policies during training and serves as a policy switching mechanism with more robustness guarantee. We first collect rollouts of $\pi_{\text{comply}}$ in simulation under random commands and external disturbances. When a failure occurs, the 150 preceding timesteps' privileged observations $\bm{o'}_{\text{unsafe}}$ in \eqref{teacherobs} which contain the observable states $\bm{s}_\text{unsafe}$ in \eqref{states} are stored to an unsafe dataset $\mathcal{D}_{\text{unsafe}}$. 
The dataset $\mathcal{D}_{\text{unsafe}}$ serves two purposes. First, during Stage II training of $\pi_{\text{safe}}$, $\bm{o'}_{\text{unsafe}}\sim\mathcal{D}_{\text{unsafe}}$ are used as state initialization, helping $\pi_{\text{safe}}$ to practice recovery from potentially dangerous conditions. Second, after $\pi_{\text{safe}}$ is trained, we initialize environment with $\bm{o'}_{\text{unsafe}}\sim\mathcal{D}_{\text{unsafe}}$ and execute $\pi_{\text{safe}}$ to determine whether it can recover the robot or not. The success and failure labels associate with each $\bm{o'}_{\text{unsafe}}$ are added to $\mathcal{D}_{\text{unsafe}}$. Finally, $V_{\text{safe}}$ is trained on $\bm{s}_{\text{unsafe}}$ and labels from $\mathcal{D}_{\text{unsafe}}$ via supervised learning using gradient descent. The data collection and training process for $V_\text{safe}$ is shown in Algotirhm \ref{alg:algo}. During deployment, the output of  $V_{\text{safe}} (\bm{s}_t) \in [0\sim1]$ will be compared to a safety threshold $\epsilon$. If output of $V_{\text{safe}}$ falls below $\epsilon$, $\pi_{\text{safe}}$ will be activated to control the robot.

\begin{algorithm}
\caption{Safety Critic Training}
\label{alg:algo}
\begin{algorithmic}[1]  
\STATE Init $\mathcal{D}_{\text{unsafe}} \gets \emptyset$ and $V_{\text{safe}}^\theta$
\WHILE{executing of $\pi_{\text{comply}}$ under disturbances}
    \IF{failure occurs at time $t$}
        \STATE $\mathcal{D}_{\text{unsafe}} \gets 
        \mathcal{D}_{\text{unsafe}} \cup \{\bm{o'}_{t-k}\}_{k=0}^{150}$
    \ENDIF
\ENDWHILE
\STATE Train $\pi_{\text{safe}}$ with $\bm{o'}_{\text{unsafe}}$ $\sim\mathcal{D}_{\text{unsafe}}$ as state initialization
\FOR{each $\bm{o'}_{\text{unsafe}}\sim \mathcal{D}_{\text{unsafe}}$}
    \STATE Initialize environment with $\bm{o'}_{\text{unsafe}}$ and execute $\pi_{\text{safe}}$ 
    \STATE $y \gets \mathbb{I}[\text{successful recovery}]$
    \STATE Assign label $y$ to $\bm{o'}_{\text{unsafe}}$ in $\mathcal{D}_{\text{unsafe}}$
\ENDFOR
\FOR{each optimization step}
    \STATE Sample batch $\big(\bm{s}_{\text{unsafe}},\, y\big) \sim \mathcal{D}_{\text{unsafe}}$
    \STATE $\mathcal{L} \leftarrow \text{BCELoss}(V_{\text{safe}}(\bm{s}_{\text{unsafe}};\theta), y)$
    \STATE $\theta \leftarrow \theta - \eta \nabla_\theta \mathcal{L}$ 
\ENDFOR
\end{algorithmic}
\end{algorithm}
\vspace{-0.9em}   

\subsection{Training details}
We train all policies in the Isaac Gym simulator \cite{makoviychuk2021isaacgymhighperformance} using 4096 environments in parallel. We use PPO algorithm \cite{schulman2017proximalpolicyoptimizationalgorithms} for policy optimization. The policy networks are parameterized as multi-layer perceptrons (MLPs) with hidden layer sizes [512,256,128]. The safety critic consists of a two-layer GRU with hidden size 256, followed by an MLP head with layers [128,64]. We employ curriculum learning by gradually increasing command velocity range and magnitude of external forces and torque. In each episode, the external disturbances are applied four times at randomly sampled timesteps with random force impact duration $T_F$. Dynamics uncertainties and sensor noise are applied to reduce the sim to real gap. The domain randomization details are shown in Table \ref{tab:domain_rand}.

\section{Simulation Experiments}
In this section, we present the results of various simulation experiments, analysis of our method and comparison with baseline methods. Specifically, we answer the following questions: \textbf{Q1} Does our switched-policy method achieve both adjustable compliance locomotion and safety under a wide range of external force disturbances? \textbf{Q2} How does the tunable parameter $\epsilon$ affect the performance? \textbf{Q3} Is the teacher-student training framework necessary and effective? \textbf{Q4} How does our method compare to other baseline methods?

\subsection{Metrics and Baselines}
We consider the following metrics to evaluate performance:

\textbf{1) Effective compliance range ($\mathrm{s/kg}$)}: $\Delta C = C_{\mathrm{high}}- C_{\mathrm{low}}$, where $C$ is the directional effective compliance based on the velocity response to external forces calculated as: 
{\small
\begin{equation}
C
=
\frac{1}{|\mathcal{T_F}|}
\sum_{t \in \mathcal{T_F}}
\frac{
\left(\bm{v}(t) - \bm{v}'(t)\right)^\top \bm{F}_{xy}(t)
}{
\bm{F}_{xy}^\top(t) \bm{F}_{xy}(t)
},
\quad
\mathcal{T_F}=\left\{t \,\middle|\, \left\|\bm{F}_{xy}(t)\right\|>0\right\},
\label{eq:compliance_level}
\end{equation}
}
which quantifies the velocity response of the robot along the direction of the applied external force. $C_{\mathrm{high}}$ and $C_{\mathrm{low}}$ correspond to the effective compliance obtained using the highest and lowest compliance parameter settings of each controller. A larger $\Delta C$ indicates a wider achievable compliance range.

\textbf{2) Command velocity tracking error ($\mathrm{m/s}$)}:
The episode-average command velocity tracking error during force-free timesteps computed as
{\small
\begin{equation}
e_v
=
\frac{1}{|\mathcal{T}|}
\sum_{t \in \mathcal{T}}
\left\|
\bm{v}(t) - \bm{v}'(t)
\right\|_2,
\quad
\mathcal{T}
=
\left\{
t \,\middle|\,
\|\bm{F}_{xy}(t)\|_2 = 0
\right\},
\label{eq:ev_noforce}
\end{equation}
}
A lower $e_v$ indicates that the robot recovers faster from external disturbance to execute the command velocity.

\textbf{3) Success rate $(\%)$}: $
SR =
\frac{N_{\mathrm{success}}}{N_{\mathrm{disturbances}}}
\times 100\%,
$
where $N_{\mathrm{success}}$ is the number of times in which the robot maintains locomotion without failure, and $N_{\mathrm{disturbances}}$ is the total number of disturbance occurrence.

\textbf{4) Average power ($\mathrm{W}$)}: The episode-average power consumed by all motors, calculated as
\begin{equation}
{P}_{\mathrm{motor}} =
\frac{1}{t_f}
\sum_{i} \int_{0}^{t_f} \left| \tau_i(t)\dot{q}_i(t) \right| \, dt.
\end{equation}

We compare our approach against two representative learning-based force compliant locomotion baseline controllers:

\textbf{1) HAC-Loco}~\cite{zhou2025hacloco}:  
A hierarchical reinforcement learning framework designed for quadruped locomotion under continuous external disturbances. We use $\beta = 10$ and $\beta = 1000$ as the highest and lowest compliance setting.

\textbf{2) FACET}~\cite{xu2025facet}:  
A reinforcement learning-based impedance tracking controller that enforces compliance by tracking the dynamics of a virtual mass–spring–damper model. We use $K_p = 0, K_d = 4$ and $K_p = 48, K_d = 13.85$ as the highest and lowest compliance setting.

We re-implemented both baselines using the same disturbance settings as our method.

\begin{table}[!t]
\vspace{5pt}
\centering
\caption{Domain Randomization Parameter}
\begin{tabular}{l l l}
\hline
\textbf{} & \textbf{Term} & \textbf{Value} \\
\hline
\multirow{6}{*}{\textbf{Dynamics}} 
    & Robot mass                  & $\mathcal{U}(12.75, 17.25)\ \mathrm{kg}$\\
    & Friction coefficient         & $\mathcal{U}(0.1, 2.0)$\\
    & Joint $K_p$ gain            & $\mathcal{U}(19, 21)$\\
    & Joint $K_d$ gain  & $\mathcal{U}(0.48, 0.52)$ \\
    & ERFI-50 \cite{campanaro2023learningdeployingrobustlocomotion} & $0.78 \mathrm{Nm} \times \mathrm{Curriculum}$\\
    & Joint bias & $\mathcal{U}(-0.08, 0.08)$\\
\hline
\multirow{4}{*}{\textbf{Sensor noises}} 
    & Base angular velocity & $\mathcal{U}(-0.25, 0.25)\ \mathrm{rad/s}$\\
    & Gravity vector         & $\mathcal{U}(-0.1, 0.1)$ \\
    & Joint positions        & $\mathcal{U}(-0.015, 0.015)\ \mathrm{rad}$\\
    & Joint velocities       & $\mathcal{U}(-1.5, 1.5)\ \mathrm{rad/s}$\\
\hline
\multirow{5}{*}{\textbf{Disturbances}} 
    & $F_{x,y}$ & $\mathcal{U}(-700, 700)\ \mathrm{N}$\\
    & $F_z$        & $\mathcal{U}(-50, 50)\ \mathrm{N}$\\
    & $\tau_{x,y}$        & $\mathcal{U}(-50, 50)\ \mathrm{Nm}$\\
    & $\tau_{z}$        & $\mathcal{U}(-20, 20)\ \mathrm{Nm}$\\
    & $T_F$ (Impact duration)    & $\mathcal{U}(0.5, 3)\ \mathrm{sec}$\\
\hline
\multirow{4}{*}{\textbf{Commands}} 
    & $k$ & $\mathcal{U}(0.0, 0.1)\ \mathrm{}$\\
    & $v_x'$        & $\mathcal{U}(-2.5, 2.5)\ \mathrm{m/s}$\\
    & $v_y'$        & $\mathcal{U}(-2, 2)\ \mathrm{m/s}$\\
    & $\omega_{z}'$        & $\mathcal{U}(-1.5, 1.5)\ \mathrm{rad/s}$\\
\hline
\end{tabular}
\label{tab:domain_rand}
\end{table}

\begin{figure}
    \centering
    \includegraphics[width=1\linewidth]{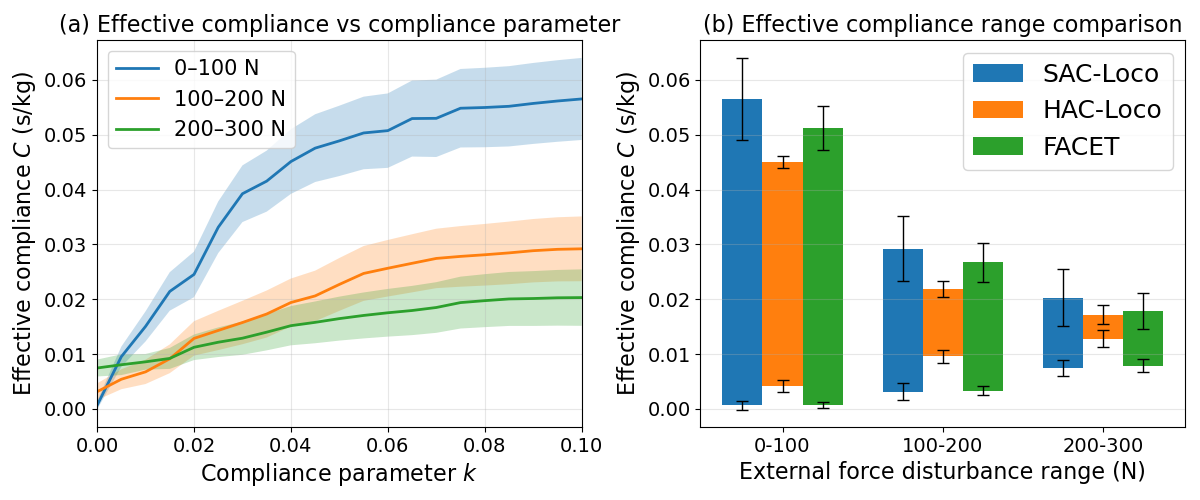}
    \caption{(a): Average effective compliance over the compliance parameters for different range of disturbance magnitude. (b): Effective compliance range of different methods.}
    \label{fig:compliance_range}
\end{figure}

\begin{table*}[t]
\vspace{5pt}
\centering
\caption{Performance under different range of external force in simulation. (Mean~$\pm$~ std dev)}
\resizebox{1\linewidth}{!}
{
\begin{tabular}{l|ccc|ccc|ccc|}
\toprule
 &$\bm{SR}$ $(\mathbf{\%})$
 & $\bm{\bar{e}_v}$ $\bm{($\mathrm{m/s}$)}$
 & $\bm{\bar{P}_{\mathrm{motor}}}$ $(\mathbf{W})$
 &$\bm{SR}$ $(\mathbf{\%})$
 & $\bm{\bar{e}_v}$ $\bm{($\mathrm{m/s}$)}$
 & $\bm{\bar{P}_{\mathrm{motor}}}$ $(\mathbf{W})$
 &$\bm{SR}$ $(\mathbf{\%})$
 & $\bm{\bar{e}_v}$ $\bm{($\mathrm{m/s}$)}$
 & $\bm{\bar{P}_{\mathrm{motor}}}$ $(\mathbf{W})$
\\
\midrule
 & \multicolumn{3}{c|}{$\bm{|F| = 0 \sim 100 \text{N}}$}
 & \multicolumn{3}{c|}{$\bm{|F| = 100 \sim 200 \text{N}}$}
 & \multicolumn{3}{c|}{$\bm{|F| = 200 \sim 300 \text{N}}$}
\\
\midrule
SAC-Loco (ours)
&$\bm{100}$
&$\bm{0.111 \pm 0.022}$
&$\bm{151.9 \pm 17.8}$&$\bm{99.84}$&$0.127 \pm 0.024$
&$\bm{160.8 \pm 21.7}$&$\bm{99.30}$&$\bm{0.146 \pm 0.024}$
&$\bm{175.8 \pm 23.2}$\\
HAC-Loco \cite{zhou2025hacloco}
&$99.74$&$0.115 \pm 0.027$&$178.2 \pm 22.3$&$94.15$&$\bm{0.121 \pm 0.029}$&$196.4 \pm 29.3$&$67.31$&${0.148 \pm 0.030}$&$217.1 \pm 27.2$\\
FACET \cite{xu2025facet}
&$99.96$&$0.158 \pm 0.041$&$165.8 \pm 19.2$&$93.65$&$0.172 \pm 0.055$&$172.3 \pm 23.8$&$80.84$&$0.195 \pm 0.062$&$189.2 \pm 24.6$\\
\midrule
 & \multicolumn{3}{c|}{$\bm{|F| = 300 \sim 400 \text{N}}$}
 & \multicolumn{3}{c|}{$\bm{|F| = 400 \sim 500 \text{N}}$}
 & \multicolumn{3}{c|}{$\bm{|F| = 500 \sim 600 \text{N}}$}
\\
\midrule
SAC-Loco (ours)
&$\bm{96.62}$&$\bm{0.183 \pm 0.041}$&$\bm{192.8\pm 26.9}$
&$\bm{91.59}$&$\bm{0.232 \pm 0.061}$&${232.5 \pm 30.8}$
&$\bm{84.11}$&$\bm{0.294 \pm 0.083}$&$255.7 \pm 31.5$
\\
HAC-Loco \cite{zhou2025hacloco}
&$43.62$&${0.196 \pm 0.038}$&$220.4 \pm 19.3$
&$28.63$&$0.275 \pm 0.052$&$225.2 \pm 20.6$
&$18.22$&${0.357 \pm 0.065}$&$231.2 \pm 18.3$
\\
FACET \cite{xu2025facet}
&$72.23$&$0.231 \pm 0.073$&$215.2 \pm 23.1$
&$56.35$&$0.306 \pm 0.071$&$\bm{223.7 \pm 28.2}$
&$35.93$&$0.368 \pm 0.080$&$\bm{228.5 \pm 28.8}$
\\
\midrule
\end{tabular}
}
\label{tab:performance}
\end{table*}

\subsection{Evaluation on compliance and safety}
We use safety threshold $\epsilon = 0.9$ for the following simulation experiments. We first evaluate the compliance performance of SAC-Loco by applying external force disturbances in three different magnitude ranges: $|\bm{F}|\in[0\sim100]$N, $|\bm{F}|\in[100\sim200]$N and $|\bm{F}|\in[200\sim300]$N. For each compliance parameter $k$, we conduct $40{,}000$ evaluation episodes with randomized commands in Table \ref{tab:domain_rand}. We compute the directional effective compliance $C$ using \eqref{eq:compliance_level} for each disturbance. As shown in Fig. \ref{fig:compliance_range} (a), the average effective compliance $C$ increases monotonically with the compliance parameter $k$, demonstrating that SAC-Loco enables adjustable compliance. For larger force ranges, the increase in $C$ becomes less pronounced due to the maximum velocity limits of the robot. Fig. \ref{fig:compliance_range} (b) compares the achievable compliance range $\Delta C$ with baseline methods. SAC-Loco achieves a larger $\Delta C$ than HAC-Loco and FACET, indicating a wider range of achievable compliance behaviors.

\begin{figure}
\centering
\includegraphics[width=1\linewidth]{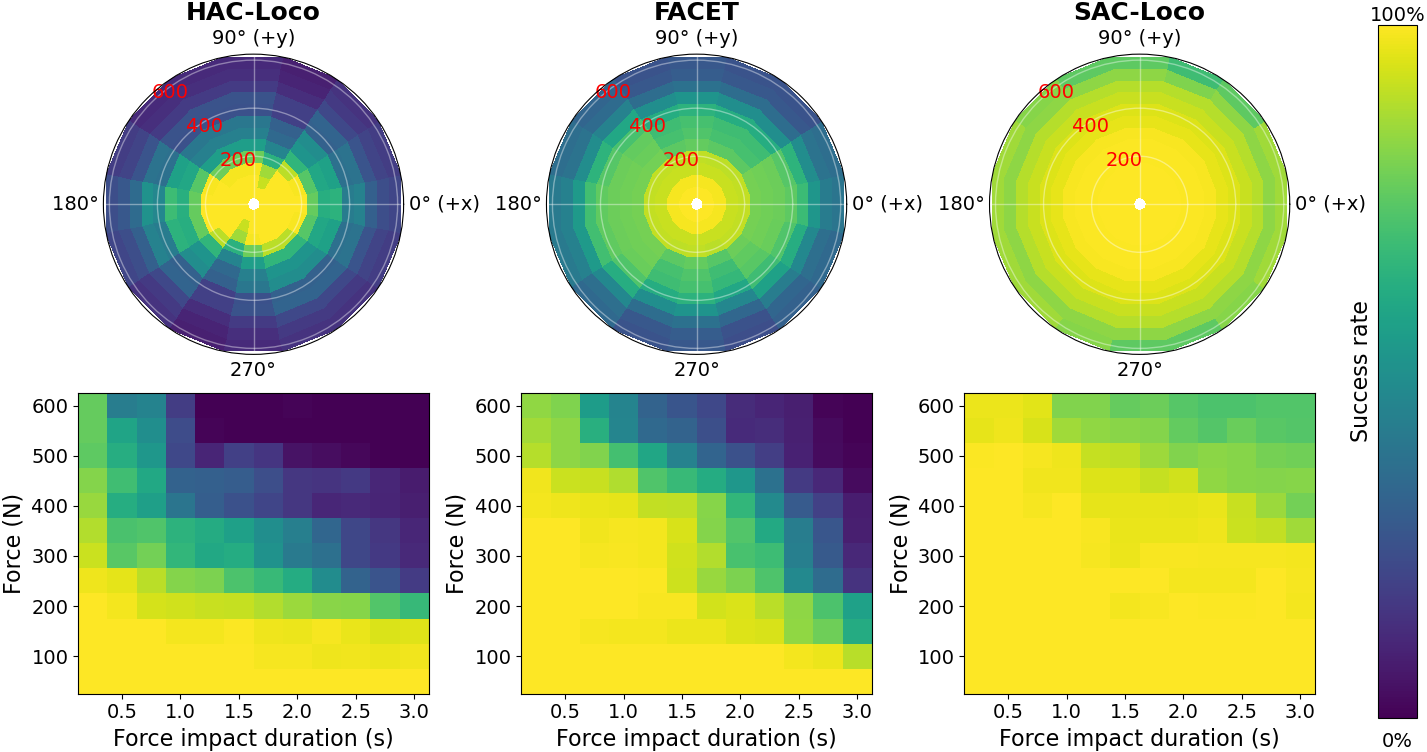}
\caption{Top: Polar heatmap showing the success rate under different force magnitude and direction of force relative to robot's heading. Bottom: Heatmap showing success rate under different force magnitude and impact duration.}
\label{fig:heatmap}
\vspace{-2.0em}  
\end{figure}
To evaluate the safety against external disturbances, we systematically test our method under different external force magnitudes $|\bm{F}|\in[50, 100, 150,..., 600]$N, impact directions relative to robot's heading $\theta_F \in [0, \pi/8, \pi/4,..., 2\pi]$ rad, and impact durations $T_F \in[0.25,0.5,...,3.0]$ s, resulting in a total of 2304 disturbance configurations. Each configuration is evaluated with 30 episodes with randomized commands in Table \ref{tab:domain_rand}. The compliance level for all methods are set to the maximum values. We calculated the $SR$ for all disturbance configurations. Fig. \ref{fig:heatmap} (top) illustrates the $SR$ averaged over all $T_F$ for each $(\bm{F}_{xy},\theta_F)$. HAC-Loco's $SR$ exhibits a noticeable decrease in $SR$ when $\bm{F}$ is applied along the lateral direction ($y$-axis) of the robot. FACET achieves overall higher $SR$ than HAC-Loco, but also has slight decrease in $SR$ under lateral disturbances. SAC-Loco outperforms both baselines and maintains a more uniform $SR$ across all $\theta_F$, demonstrating improved robustness to external forces. Fig. \ref{fig:heatmap} (bottom) illustrates the $SR$ averaged over all $\theta_F$ for each $(\bm{F}_{xy},T_F)$. HAC-Loco achieves good $SR$ with low $|\bm{F}|$ but more failures with larger $|\bm{F}|$. FACET achieves better $SR$ for larger $|\bm{F}|$ but still fails under longer $T_F$. SAC-Loco maintains higher $SR$ under both larger $|\bm{F}|$ and longer $T_F$ than baselines.
Table \ref{tab:performance} summarizes the performance. It is shown that SAC-Loco consistently achieves higher $SR$ than baselines. SAC-Loco consumes less power than baselines except when $F$ exceeds $400\text{N}$. SAC-Loco also achieves lower $e_v$, suggesting it restores the robot from disturbance compliance to velocity tracking faster. The results show that SAC-Loco achieves better safety without sacrificing on command tracking.  

\subsection{Effect of different $\epsilon$ for policy switching}
\begin{figure}
    \centering
    \includegraphics[width=0.76\linewidth]{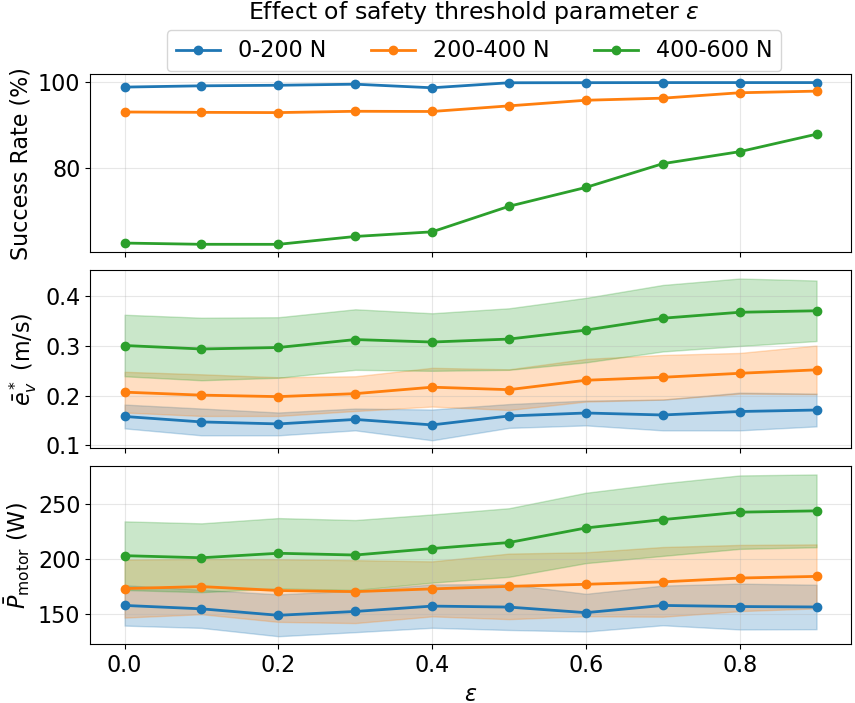}
    \caption{The effect of the safety critic output threshold $\epsilon$ on SAC-Loco's performance under different force disturbance ranges.}
    \label{fig:epislon}
    \vspace{-0.7em}  
\end{figure}

$\epsilon$ determines the threshold for policy-switching. To investigate the sensitivity of $\epsilon$ and how switching to $\pi_\text{safe}$ affect performance, we perform simulation experiments across different values of $\epsilon \in [0, 0.1, 0.2, ..., 0.9]$. For each $\epsilon$, we conduct $40{,}000$ evaluation episodes and calculate the $SR$ and $\bar{P}_{\mathrm{motor}}$. In addition, we calculate the velocity tracking error between robot's velocity and the desired velocity $\bm{v}^*$ generated from the velocity modulator as ${e}^*_v
=
\frac{1}{T}
\sum_{t=0}^{T}
\left\|
\bm{v}(t)
-
\bm{v}^*(t)
\right\|_2$, a larger ${e}^*_v$ indicates more deviation from the compliance behavior of $\pi_\text{comply}$. Fig. \ref{fig:epislon} shows that when $|\bm{F}|<200$N, the performance is insensitive to changes in $\epsilon$, indicating that $\pi_\text{safe}$ is rarely activated under moderate disturbance. This suggests that the output of $V_{\text{safe}}$ accurately characterizes the safety condition of the robot. For $|\bm{F}|>200$N, setting $\epsilon = 0$ (using only $\pi_\text{comply}$) leads to a decrease in $SR$. As $\epsilon$ increases, $SR$ improves, indicating that switching to $\pi_\text{safe}$ helps the robot recover to safe states. Meanwhile, increasing $\epsilon$ results in only marginal increase in ${e}^*_v$ and ${P}_{\mathrm{motor}}$, indicating that our method improves safety without significantly compromising compliance and energy usage. 

\subsection{Policy distillation performance}
\begin{figure}
    \centering
    \includegraphics[width=0.9\linewidth]{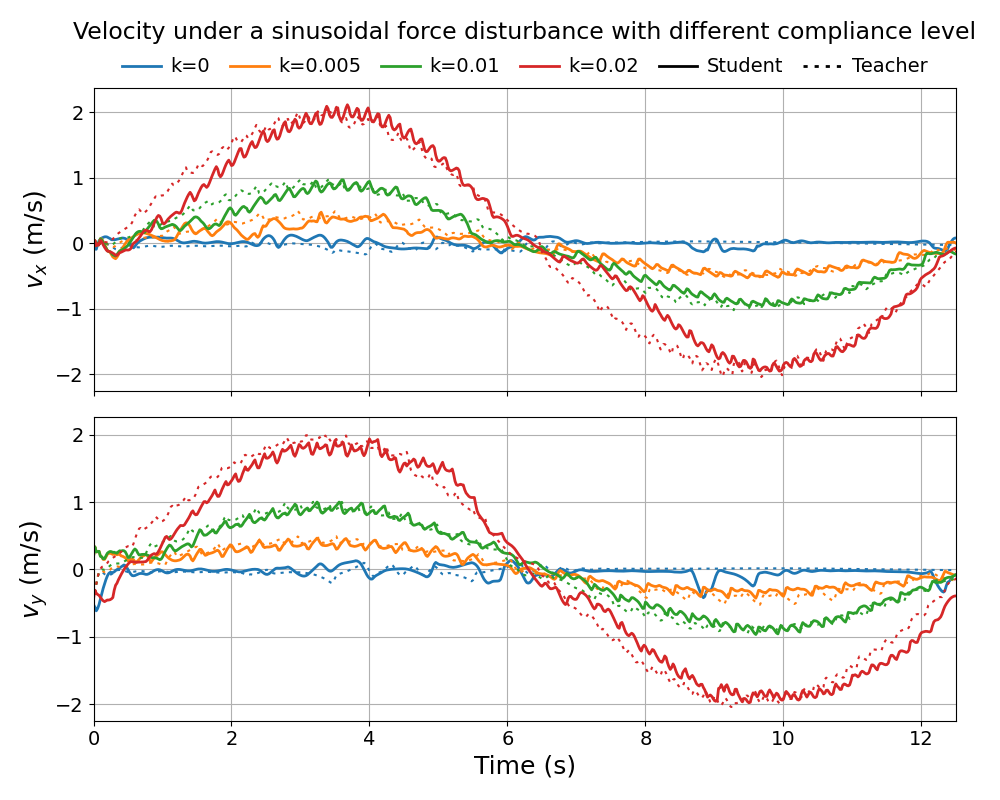}
    \caption{Velocity of the robot under a sinusoidal force disturbance: $F_x = F_y = 100\text{sin}(0.5t)$ with different compliance parameters. Dashed line is teacher and solid line is student.}
    \label{fig:teacher_student}
\end{figure}


To verify that our teacher-student framework helps the robot to learn external disturbance with only proprioceptive sensing, we compare the performance of teacher policies with performance of student policies to verify the effectiveness of policy distillation. Fig. \ref{fig:teacher_student} shows that $\pi_\text{comply}$ successfully achieves different compliance levels under a sinusoidal force, tracking $\bm{v}^*$ with small tracking errors as compared to $\pi_\text{comply}^*$. We further conduct $40,000$ episodes and calculated performance metrics using the teacher's policies. Table \ref{tab:ablations} shows that the student policy performs slightly below the teacher policy. This result validates that the distillation framework successfully preserves compliance and recovery capabilities while removing reliance on privileged force information, making the policy suitable for real-world deployment.


\subsection{Ablation studies on SAC-Loco}

\begin{table}[t]
\vspace{5pt}
\centering
\caption{Ablation on SAC-Loco}
\resizebox{0.95\linewidth}{!}
{
\begin{tabular}{l|ccc}
\toprule

 & $\bm{SR}$ $(\%)$
 & $\bm{\bar{e}^*_v}$ $(\mathrm{m/s})$& $\bm{\bar{P}_{\mathrm{motor}}}$ $(\mathrm{W})$
\\

\midrule
\multicolumn{4}{c}{$|\bm{F}| = 0 \sim 300\mathrm{N}$}
\\
\midrule

SAC-Loco (Teacher)
& $99.78$& ${0.171 \pm 0.011}$& ${158.7 \pm 14.2}$\\

\textbf{SAC-Loco (Ours)}
& $\bm{99.71}$& $\bm{0.190 \pm 0.035}$& $\bm{162.8 \pm 23.2}$\\

SAC-Loco w/o teacher
& $70.21$& $0.719 \pm 0.253$& $221.2 \pm 25.8$\\

$\pi_\text{safe}$ w/o $\bm{o'}_{\text{unsafe}}$
& $99.36$& $0.198 \pm 0.041$
& $168.8 \pm 15.2$\\

$\pi_\text{safe}$ w/o CCP
& $92.13$& $0.194 \pm 0.033$& $183.8 \pm 22.4$\\
SAC-Loco w/o $V_{\text{safe}}$
& $99.42$& $0.207 \pm 0.090$& $195.8 \pm 39.2$
\\

\midrule
\multicolumn{4}{c}{$|\bm{F}| = 300 \sim 600\mathrm{N}$}
\\
\midrule

SAC-Loco (Teacher)
& $93.81$& ${0.252 \pm 0.023}$& ${219.6\pm 20.1}$\\

\textbf{SAC-Loco (Ours)}
& $\bm{90.77}$& $\bm{0.285 \pm 0.092}$& $\bm{227.0\pm 39.5}$\\

SAC-Loco w/o teacher
& $21.19$& $1.283 \pm 0.552$& $275.8 \pm 31.4$\\

$\pi_\text{safe}$ w/o $\bm{o'}_{\text{unsafe}}$
& $85.21$& $0.288 \pm 0.085$& $211.4 \pm 30.6$\\

$\pi_\text{safe}$ w/o CCP
& $30.96$& $0.454 \pm 0.140$& $262.9 \pm 36.0$\\
SAC-Loco w/o $V_{\text{safe}}$
& $68.38$& $0.659 \pm 0.189$& $237.1 \pm 37.7$\\

\bottomrule

\end{tabular}
}
\label{tab:ablations}
\end{table}

We perform the following ablation studies:

\textbf{1) SAC-Loco w/o teacher}: We train $\pi_\text{comply}$ and $\pi_\text{safe}$ directly using student observations as \eqref{studentobs} with the same training setup. 

\textbf{2) $\bm{\pi}_\text{safe}$ w/o $\bm{o'}_{\text{unsafe}}$}: We train $\pi_\text{safe}$ without using $\bm{o'}_{\text{unsafe}}$ as state initializations.

\textbf{3) $\bm{\pi}_\text{safe}$ w/o CCP}: We train $\pi_\text{safe}$ without the two stage training and without using the corrected capture point as recovery heuristic. We directly use reward to penalize collision and failure.

\textbf{4) SAC-Loco w/o $V_{\text{safe}}$}: We replace $V_{\text{safe}}$ with a rule-based policy switching state machine such that the system switches to $\pi_\text{safe}$ when $\|{\omega}_{xy}\|_2 > 1$ rad/s.

As shown in Table \ref{tab:ablations}. The teacher-student training framework is necessary. Without the teacher's guidance, $\pi_\text{comply}$ cannot track $\bm{v}^*$ and $\pi_\text{safe}$ is unable to recover the robot due to unobservable $\bm{F}$ and $\bm{v^*}$, resulting in more failure, worse velocity tracking and more power usage. Removing $\bm{o'}_{\text{unsafe}}$ as state initialization slightly decreases $SR$. Training $\pi_\text{safe}$ without CCP dynamics drastically decrease $SR$, indicating the effectiveness of using CCP to recover robot from unsafe states. Using rule based policy switching also results in more failure, highlighting the effectivess of the safety critic.

\section{Hardware deployment}

\begin{figure}
    \centering
    \includegraphics[width=1\linewidth]{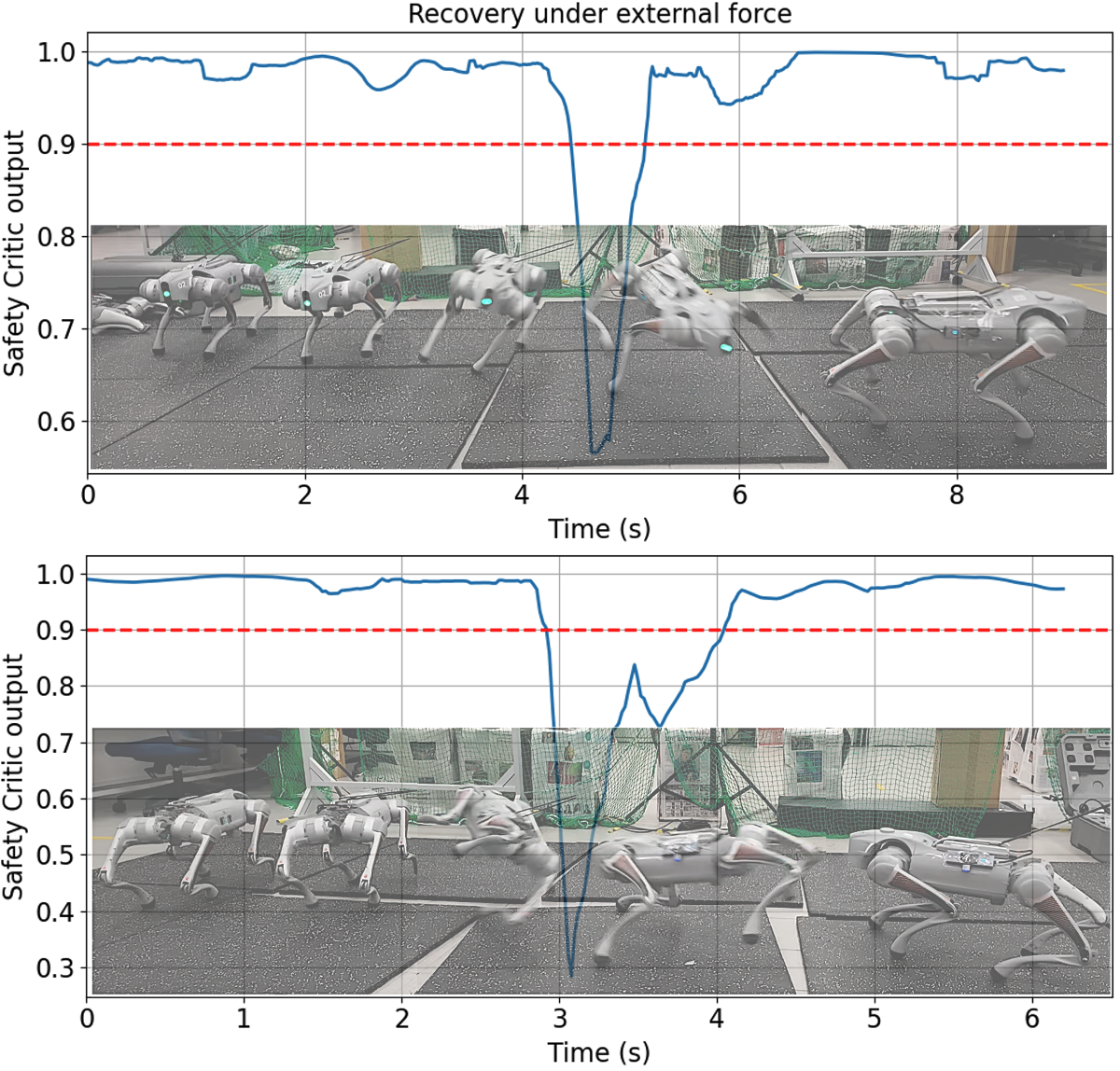}
    \caption{Two example cases when the robot being dragged and losing balance. $V_\text{safe}$ outputs a value below $\epsilon = 0.9$ and $\pi_\text{safe}$ takes over to restore balance. After output of $V_\text{safe}$ increases above $0.9$, $\pi_\text{comply}$ is switched back. In both cases, robot restores balance and adjust the heading to align its $x$-axis with the force direction.}
    \label{fig:recover_robot}
\end{figure}

\begin{figure*}
\vspace{8pt}
    \centering
    \includegraphics[width=0.95\linewidth]{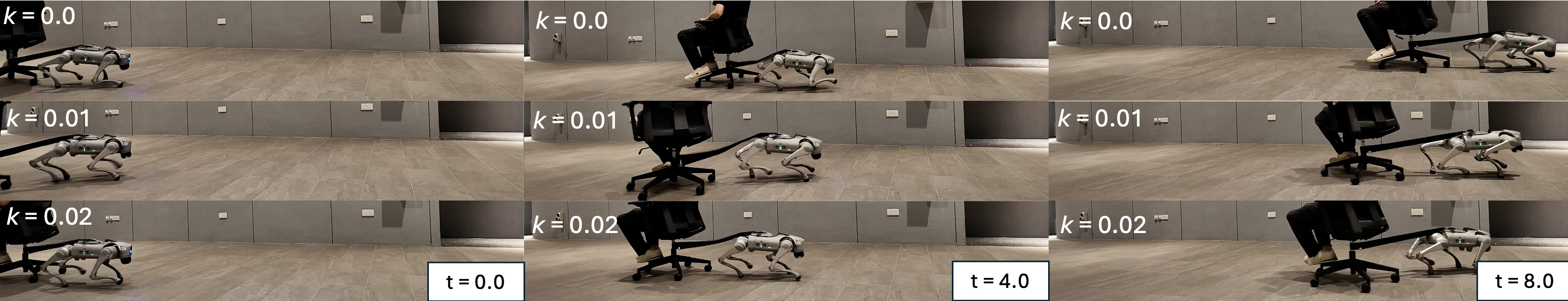}
    \caption{Quadruped robot is connected to an office chair with a seated person, and the robot is set to a constant velocity command of $v_x' =0.6$ m/s to pull the chair forward. By adjusting different values of $k$, the pulling speed is varied.}
    \label{fig:chair}
    \vspace{-0.7em}  
\end{figure*}

We deploy SAC-Loco onto the Unitree Go2 quadruped robot. The policy runs at 50 Hz on the Unitree Go2 Jetson Orin NX. We perform experiments to verify our policy's effectiveness. 

First, we test the compliance by using another quadruped robot to pull the robot with our policy. Given different compliance parameter, the robot can either resist the pulling force or comply with the force. We also tied the robot to an office chair with a person in the seat. The robot is given a constant velocity command of $v_x' =0.6m/s$. The total weight of chair and person is around 70 kg. As shown in Fig. \ref{fig:chair}, the robot's pulls the chair with lower speed as $k$ increases. 

Second, we evaluate safety under large external disturbances by dragging the robot using a rope connected to a digital force gauge. The robot is pulled by a person with the intent of inducing a failure, while the digital force gauge records the peak force $F_{\text{peak}}$ applied during each pull. We collect $5$ failures for each baseline. Among the $5$ failures, HAC-Loco experiences an average $F_{\text{peak}} = 120.4$ N. FACET experiences an average $F_{\text{peak}} = 193.8$ N. In contrast, SAC-Loco has $0$ failures in this experiment. Fig. \ref{fig:recover_robot} shows two example cases where the robot switches from $\pi_\text{comply}$ to $\pi_\text{safe}$ promptly to restore balance during an external force disturbance. Fig. \ref{fig:recover_robot} (top) shows when force is toward the head of the robot, the robot turns to align its head with the force. In contrast, Fig. \ref{fig:recover_robot} (bottom) shows that when the force is toward the tail of the robot, the robot turns to align its tail with the force.

Third, we set the maximum velocity command $v_x' = 2.5$ m/s with compliance parameter $k = 0$. We pull the robot using rope with the digital force gauge in the opposite direction and measure the largest pulling force that the robot can exert while maintaining stable locomotion without failure. The largest forward and backward pulling forces are $10.54$ kg and $10.45$ kg, respectively. This outperforms the $7.5$ kg and $10$ kg forward and backward pulling force reported in FACET \cite{xu2025facet}.


\section{Conclusion}
This paper presented SAC-Loco, a novel safety-aware force compliant controller for quadruped robots to achieve adjustable compliant velocity tracking under external disturbances and robustness against large impulses. Extensive simulation and hardware experiments validate the effectiveness and practicality of SAC-Loco. Compared with previous work, our approach broadens the compliance and safety of a quadruped under wider range of external forces. This work opens avenues for future research including adaptive adjustment of compliance level in response to varying environments and integration of the safety mechanism into highly dynamic quadruped maneuvers.


\bibliographystyle{IEEEtran}
\bibliography{reference.bib}

\end{document}